\documentclass[conference]{IEEEtran}
\IEEEoverridecommandlockouts
\usepackage{cite}
\usepackage{lettrine} 
\usepackage{lipsum}
\usepackage[utf8]{inputenc}
\usepackage{graphicx}    
\usepackage{subcaption}
\usepackage{tikz}
\usetikzlibrary{positioning,arrows.meta,calc,fit}
\usepackage{multirow}
\usepackage{booktabs}
\usepackage{amsmath,amssymb,amsfonts}
\usepackage{algorithmic}
\usepackage{graphicx}
\usepackage{textcomp}
\usepackage{xcolor}
\usepackage{amsmath}
\usepackage{url}
\usepackage{float}
\usepackage{graphicx}
\usepackage{hyperref}
\usepackage{hyperref}  
\usepackage{mathrsfs}
\usepackage[utf8]{inputenc}
\usepackage{mathrsfs}
\usepackage{amsmath,amssymb,amsfonts}
\usepackage{tabularx,booktabs,makecell,caption}
\usepackage{caption}
\usepackage{eso-pic}

\def\BibTeX{{\rm B\kern-.05em{\sc i\kern-.025em b}\kern-.08em
    T\kern-.1667em\lower.7ex\hbox{E}\kern-.125emX}}
\begin{document}

\title{\vspace*{1cm}Recurrent Deep Reinforcement Learning for Chemotherapy Control under Partial Observability 
}


\makeatletter
\newcommand{\linebreakand}{%
  \end{@IEEEauthorhalign}
  \hfill\mbox{}\par
  \mbox{}\hfill\begin{@IEEEauthorhalign}
}
\makeatother
\IEEEoverridecommandlockouts
\author{
\IEEEauthorblockN{1\textsuperscript{st} Firas Mohamed Elamine Kiram}
\IEEEauthorblockA{\textit{Dept. of Information Engineering} \\
\textit{University of Padova} \\
Padova, Italy \\
\textit{Dept. of Computer Science} \\
\textit{University of Biskra, Algeria} \\ 
kiramfiras@dei.unipd.it}
\and
\IEEEauthorblockN{2\textsuperscript{nd} Imane Youkana}
\IEEEauthorblockA{\textit{LINFI Lab, Dept. of Computer Science} \\
\textit{University of Biskra}\\
Biskra, Algeria \\
imane.youkana@univ-biskra.dz}
\and
\IEEEauthorblockN{3\textsuperscript{rd} Rachida Saouli}
\IEEEauthorblockA{\textit{LINFI Lab, Dept. of Computer Science} \\
\textit{University of Biskra}\\
Biskra, Algeria \\
rachida.saouli@univ-biskra.dz}
\linebreakand
\IEEEauthorblockN{4\textsuperscript{th} Gian Antonio Susto}
\IEEEauthorblockA{\textit{Dept. of Information Engineering} \\
\textit{University of Padova}\\
Padova, Italy \\
gianantonio.susto@unipd.it}
\and
\IEEEauthorblockN{5\textsuperscript{th} Laid Kahloul}
\IEEEauthorblockA{\textit{LINFI Lab, Dept. of Computer Science} \\
\textit{University of Biskra}\\
Biskra, Algeria \\
l.kahloul@univ-biskra.dz}
}
\maketitle

\begin{abstract}
Chemotherapy dose optimization can be formulated as a dynamic treatment regime, requiring sequential decisions under uncertainty that must balance tumor suppression against toxicity. However, most reinforcement learning approaches assume full observability of the patient state, a condition rarely met in clinical practice. We investigate whether memory-augmented policies can improve chemotherapy control under partial observability. To this end, we employ a recurrent TD3-based approach with separate LSTM actor–critic networks and evaluate it on the \textit{AhnChemoEnv} benchmark from DTR-Bench, considering both off-policy and on-policy recurrent architectures against feed-forward TD3 and Soft Actor–Critic. Pharmacokinetic and pharmacodynamic variability are held fixed to isolate hidden-state uncertainty and observation noise and to avoid confounding effects from inter-patient variability. Across ten random seeds, recurrence yields modest benefit under full observability but substantially stronger and more stable performance under partial observability, with more consistent tumor suppression and improved normal-cell preservation. These findings indicate that memory-based policies are particularly beneficial when clinically relevant state information is incomplete or noisy.

\end{abstract}

\begin{IEEEkeywords}
Chemotherapy dose optimization,  dynamic treatment regime, reinforcement learning, partial observability, recurrent architectures, actor--critic methods.
\end{IEEEkeywords}

\renewcommand{\LettrineFontHook}{\bfseries\scshape} 
\setcounter{DefaultLines}{2} 
\setlength{\DefaultFindent}{2pt} 
\setlength{\DefaultNindent}{0pt} 

\section{Introduction}
Cancer remains a significant public health \mbox{concern.} \mbox{The American Cancer Society} (ACS) estimates that about 2.1 million new cancer cases will be diagnosed in the United States in 2026~\cite{ACS2026}. In chemotherapy, treatment decisions must balance tumor suppression against toxicity and damage to healthy or immune cells. However, drug dose and treatment schedule are typically guided by standard protocols, which are often population-based and only weakly adaptive, limiting their ability to capture inter-patient variability and the patient’s evolving condition during the treatment.

Consequently, the limitations of such protocol-based \mbox{approaches} have increasingly been recognized in both scientific and clinical communities \cite{OPTIMALCONTROLLIMITATIONS}. Personalized medicine, which aims to tailor treatment to each patient based on individual characteristics \cite{PersonalizedMedicineProgressandPromise}, has been a major goal in healthcare. 
Within this broader paradigm, Dynamic Treatment Regimes (DTRs) have emerged as an important component of personalized medicine\cite{DTR,DTR2}. DTRs consist of a sequence of individualized decision rules designed to optimize long-term patient outcomes by adapting treatment over time according to patient characteristics and evolving responses. As such, they provide a principled framework for systematically guiding treatment decisions and delivering personalized care.

Given the inherent nonlinearity and uncertainty of cancer dynamics, as well as the need for long-term sequential treatment decisions, reinforcement learning (RL) \cite{sutton_barto_rl_2018} provides a promising methodological framework. RL is well suited to sequential decision-making problems \cite{Deliu2024RLPopulation}, in which an agent interacts with an environment and learns actions that maximize cumulative reward over time. In this setting, treatment policies can be learned directly through interaction with the environment, without requiring explicit knowledge of the patient's internal dynamics or a detailed mathematical model from which to derive an optimal dosing strategy. Deep reinforcement learning (DRL) has demonstrated strong performance in high-dimensional continuous-control tasks \cite{DDPGforcontinuouscontrol,DRLforchemotherapy}. By operating directly in continuous state and action spaces, it avoids the expert-defined discretization required in earlier reinforcement learning approaches \cite{Padmanabhan2017RLchemo,MartinGuerrero2009RLdosing}. This property is particularly relevant for medical decision-making problems such as chemotherapy dosing, where treatment actions naturally lie in a continuous range. Consequently, DRL enables the development of more realistic, flexible, and adaptive closed-loop treatment strategies aligned with the principles of precision oncology.

Recent benchmark efforts have provided standardized \mbox{simulation} environments for evaluating reinforcement learning methods in dynamic treatment regime problems, including chemotherapy treatment optimization \cite{DTRBench}. A fundamental challenge in real clinical settings is that the patient's full internal physiological state is not directly observable, and treatment decisions must therefore be made using only partial clinical information\cite{Liang2025DTR}. Motivated by this challenge, we study chemotherapy treatment optimization under partial observability. Prior work applied TD3 to chemotherapy dose control on the same benchmark under full observability~\cite{Mashayekhi2024DRLChemo}; however, the partially observable setting, where clinically relevant state variables are hidden, has not been systematically investigated with recurrent actor--critic methods. Unlike benchmark settings that also consider \mbox{pharmacokinetic/pharmacodynamic} (PK/PD) variability, we keep PK/PD parameters fixed so that the role of the recurrence can be examined more directly under partial observability and noise, without the added effect of inter-patient heterogeneity. The main contributions of this paper are as follows:
\begin{itemize}
    \item We evaluate recurrent and non-recurrent deep RL on the \textit{AhnChemoEnv} POMDP benchmark with fixed PK/PD parameters, isolating the effect of partial observability from inter-patient variability.
    \item We present a systematic comparison of recurrent and memoryless policies under both fully and partially observable conditions, extending prior full-observability evaluations~\cite{Mashayekhi2024DRLChemo}.
    \item We demonstrate empirically that memory-based policies are substantially more robust to hidden-state uncertainty and observation noise, reducing cross-seed variance by more than an order of magnitude relative to memoryless baselines.
    \item We provide trajectory-level analysis showing that recurrent policies yield more consistent tumor suppression, stronger normal-cell recovery, and more structured dosing behavior.
\end{itemize}

The remainder of this paper is organized as follows. Section~II describes the POMDP formulation, the ODE-based cancer model, and the recurrent TD3 architecture. Section~III presents the experimental results, and Section~IV concludes the paper with limitations and future directions.

\section{Methods}

\subsection{POMDP Formulation for Chemotherapy Optimization}

Chemotherapy treatment optimization can be naturally formulated as a partially observable Markov decision process (POMDP), since the full physiological state is not directly observable during treatment \cite{AnInverseRLforPOMDPHealthcare,PlanningActingPOMDP}. A POMDP is defined by the tuple $(\mathcal{S}, \mathcal{A}, \mathcal{P}, \Omega, \mathcal{O}, \mathcal{R}, \gamma)$. The latent state space $\mathcal{S}$ represents the underlying physiological condition. In our setting, the latent state at time $t$
is given by \mbox{\(s_t = [N(t), T(t), I(t), B(t)]^\top\)}, where $N(t)$, $T(t)$, $I(t)$, and $B(t)$ denote the populations of normal cells, tumor cells, immune cells, and the drug concentration in the bloodstream, respectively. The action space $\mathcal{A}$ corresponds to chemotherapy dosing decisions, where the agent selects the drug dose at each decision step. The transition dynamics are described by $\mathcal{P}(s_{t+1}\mid s_t,a_t)$, which defines the probability of transitioning from state $s_t$ to state $s_{t+1}$ after applying action $a_t$. The observation space $\Omega$ denotes the set of observable clinical variables available to the agent, and the observation $o_t \in \Omega$ is generated according to the observation model $\mathcal{O}(o_t \mid s_t)$. In the partially observable setting considered in this work, the normal-cell population is hidden from the agent. Thus, the observable part of the state is given by \mbox{\(g(s_t) = [T(t), I(t), B(t)]^\top\)}. Each observed component is then corrupted by independent multiplicative uniform noise: 
\begin{equation}
o_t = g(s_t) + \alpha_{\text{obs}}\, g(s_t)\odot \xi_t,
\qquad \xi_{t,i} \sim \mathcal{U}(-0.5,0.5),
\end{equation}
where \(o_t\) is the observation, \(\xi_t\) is a noise vector, \(\alpha_{\mathrm{obs}}\) is the observation-noise level, and $\odot$ denotes the Hadamard product. In addition, multiplicative Gaussian noise is applied to each ordinary differential equation (ODE) derivative:
\begin{equation}
\dot{s}_{t,i}^{\text{noisy}} = \dot{s}_{t,i}(1+\epsilon_{t,i}),
\qquad \epsilon_{t,i} \sim \mathcal{N}(0,\alpha_{\text{state}}),
\end{equation}
where \(\dot{s}_{t,i}\) is the deterministic state derivative, \(\dot{s}^{\mathrm{noisy}}_{t,i}\) its noisy counterpart, \(\epsilon_{t,i}\) a Gaussian perturbation term, and \(\alpha_{\mathrm{state}}\) the state-noise level. Treatment decisions must therefore be made from incomplete and noisy clinical information. The reward function $\mathcal{R}$ follows the released benchmark environment used in our experiments. Specifically, the step reward is given by 
\begin{equation}
r_t = \mathcal{R}(s_t, a_t, s_{t+1}) = \frac{N_t}{N_0} - \frac{T_t}{T_0} - u_t,
\end{equation}
which rewards preservation of normal cells, penalizes tumor burden, and discourages excessive drug administration. A terminal penalty is applied when the episode ends due to violation of the normal-cell safety threshold.
The discount factor $\gamma \in [0,1]$ controls the relative importance of immediate and future outcomes. The objective is to learn a policy based on the available observation history that maximizes the expected discounted return over a treatment episode:
\[
J(\pi) = \mathbb{E}_{\tau \sim \pi}\left[\sum_{t=0}^{H-1} \gamma^t r_t\right].
\]

In practice, the true transition dynamics are unknown. Therefore, policy learning is performed \emph{in silico} using a simulator derived from mathematical models of tumor growth, immune response, and chemotherapy effects, which serves as a surrogate environment for evaluating treatment strategies\cite{DTRBench}.

\subsection{Mathematical Model of Cancer Chemotherapy}

To model the dynamics of chemotherapy treatment, we adopt the four-state nonlinear ordinary differential equation (ODE) system used in the \textit{AhnChemoEnv} benchmark environment from DTR-Bench \cite{DTRBench}, which is derived from prior mathematical models of tumor--immune interactions and chemotherapy response \cite{ODE1,ODE2}. The model captures the coupled evolution of normal cells \(N(t)\), tumor cells \(T(t)\), immune cells \(I(t)\), and the chemotherapeutic drug concentration in the bloodstream \(B(t)\). The control input \(u(t)\) represents the administered chemotherapy dosage. This model provides a compact mechanistic description of the interaction among tumor growth, healthy tissue, immune response, and chemotherapy exposure, making it suitable for \textit{in silico} treatment optimization. The dynamics are given by:

\begin{equation}
\label{eq:system_ODEs}
\left\{
\begin{aligned}
    \frac{dN}{dt} &= r_2 N(1 - b_2 N) - c_4 T N - a_3(1 - e^{-B})N, \\
    \frac{dT}{dt} &= r_1 T(1 - b_1 T) - c_2 I T - c_3 T N - a_2(1 - e^{-B})T, \\
    \frac{dI}{dt} &= s + \frac{\rho I T}{\alpha + T} - c_1 I T - d_1 I - a_1(1 - e^{-B})I, \\
    \frac{dB}{dt} &= -d_2 B + u(t)
\end{aligned}
\right.
\end{equation}

\vspace{1em}
The ODE parameter values used in this work follow the benchmark implementation of \textit{AhnChemoEnv} in DTR-Bench \cite{DTRBench} and are consistent with prior chemotherapy-control studies \cite{Mashayekhi2024DRLChemo, Padmanabhan2017RLchemo}. 
In the benchmark implementation used in this work, an episode is terminated when the normal-cell population falls below 70\% of its initial value $N_0$, corresponding to a critical safety condition.


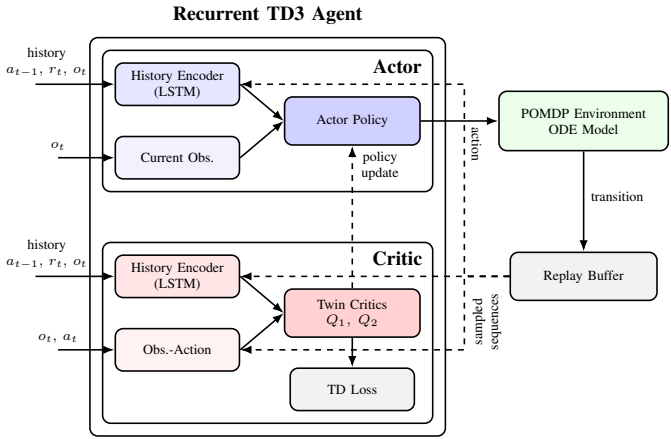
\begin{figure}[H]
\centering
\resizebox{\columnwidth}{!}{%
\begin{tikzpicture}[
    font=\scriptsize,
    >={Latex[length=2mm]},
    box/.style={
        draw, rounded corners, thick, align=center,
        minimum width=2.4cm, minimum height=0.85cm
    },
    sbox/.style={
        draw, rounded corners, thick, align=center,
        minimum width=2.2cm, minimum height=0.75cm
    },
    group/.style={draw, rounded corners, thick, inner sep=6pt},
    flow/.style={->, thick},
    train/.style={->, thick, dashed}
]

\node[sbox, fill=blue!10] (ahist) at (0,2.4) {History Encoder\\(LSTM)};
\node[sbox, fill=blue!5]  (aobs)  at (0,1.1) {Current Obs.};
\node[box,  fill=blue!18] (actor) at (3.1,1.75) {Actor Policy};

\draw[flow] (ahist.east) -- (actor.west);
\draw[flow] (aobs.east)  -- (actor.west);

\node[group, fit=(ahist)(aobs)(actor), label={[font=\bfseries,yshift=-5.5mm,xshift=23mm]above:Actor}] (agroup) {};

\draw[flow] ($(ahist.west)+(-1.4,0)$) -- (ahist.west)
    node[midway, above, align=center, xshift=-5mm] {history\\$a_{t-1},\,r_t,\,o_t$};
\draw[flow] ($(aobs.west)+(-1.0,0)$) -- (aobs.west)
    node[midway, above, xshift=-5mm] {$o_t$};

\node[sbox, fill=red!10]  (chist)  at (0,-1.0) {History Encoder\\(LSTM)};
\node[sbox, fill=red!5]   (coa)    at (0,-2.3) {Obs.-Action};
\node[box,  fill=red!18]  (critic) at (3.1,-1.65) {Twin Critics\\$Q_1,\;Q_2$};
\node[sbox, fill=gray!10] (tdloss) at (3.1,-3.0) {TD Loss};

\draw[flow] (chist.east) -- (critic.west);
\draw[flow] (coa.east)   -- (critic.west);
\draw[flow] (critic.south) -- (tdloss.north);

\node[group, fit=(chist)(coa)(critic)(tdloss), label={[font=\bfseries,yshift=-5.5mm,xshift=23mm]above:Critic}] (cgroup) {};

\draw[flow] ($(chist.west)+(-1.4,0)$) -- (chist.west)
    node[midway, above, xshift=-5mm, align=center] {history\\$a_{t-1},\,r_t,\,o_t$};
\draw[flow] ($(coa.west)+(-1.0,0)$) -- (coa.west)
    node[midway, above, xshift=-5mm] {$o_t,\,a_t$};

\node[box, minimum width=2.6cm, fill=gray!10] (buffer) at (7.2,-1.0) {Replay Buffer};
\node[box, minimum width=3.0cm, minimum height=1.05cm, fill=green!8] (env) at (7.2,1.75) {POMDP Environment\\ODE Model};

\draw[flow] (actor.east) -- node[rotate=-90, xshift=4mm, yshift=3mm] {action} (env.west);
\draw[flow] (env.south) -- ++(0,-0.55) -| node[pos=0.60,right] {transition} (buffer.north);

\draw[train] (buffer.west) -- ++(-0.8,0) |- node[pos=0.25, rotate=90, xshift=-8mm, yshift=-4mm, align = center] {sampled \\ sequences} (chist.east);
\draw[train] (buffer.west) -- ++(-0.8,0) |- (coa.east);
\draw[train] (buffer.west) -- ++(-0.8,0) |- (ahist.east);

\draw[train] (critic.north) -- ++(0,0.55) -| node[pos=0.83,above,align=center,xshift=14,yshift=-2] {policy \\ update} (actor.south);

\node[group, fit=(agroup)(cgroup), label={[font=\bfseries,yshift=1mm]above:Recurrent TD3 Agent}] {};

\end{tikzpicture}%

}
\caption{Recurrent TD3 architecture adapted from \cite{ni2022recurrentmodelfreerlstrong} for the POMDP chemotherapy task.}
\label{fig:rtd3_clean_ieee}
\end{figure}

\subsection{Recurrent TD3 under Partial Observability}
\begin{figure*}[!t]
    \centering
    \includegraphics[width=\textwidth]{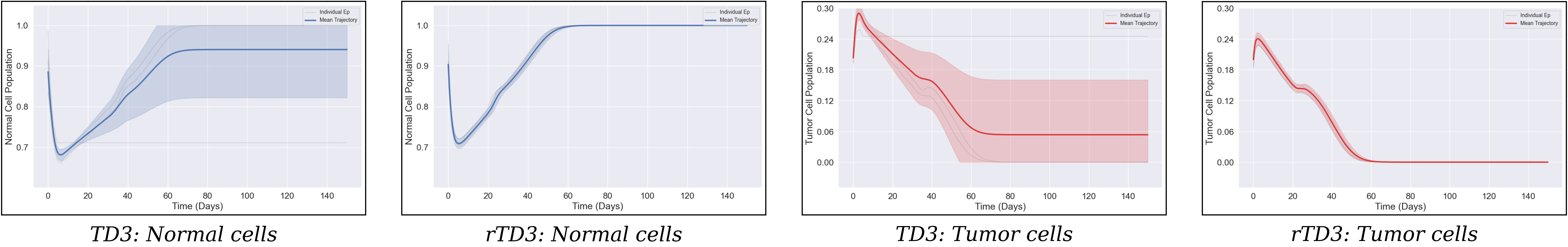}
    \caption{Trajectory-level evaluation under partial observability over 30 \textit{in silico} episodes. rTD3 yields more consistent tumor suppression and better normal-cell recovery than TD3.}
    \label{fig:traj_normal_tumor}
\end{figure*}


To address partial observability in the chemotherapy control problem, we adopt a recurrent variant of Twin Delayed Deep Deterministic Policy Gradient (TD3)\cite{TD3OriginalPaper}. In our setting, the normal-cell population is hidden and the available measurements are noisy, making the current observation alone insufficient. Following \cite{ni2022recurrentmodelfreerlstrong}, we apply the recurrent TD3 framework to our chemotherapy POMDP. This design allows the policy to exploit temporal information from past interactions and infer hidden aspects of the underlying system. The original study also found that recurrent \mbox{actor--critic} design choices and context length play an important role in achieving strong performance under partial observability.

\begin{figure}[t]
    \centering
    \includegraphics[width=\linewidth]{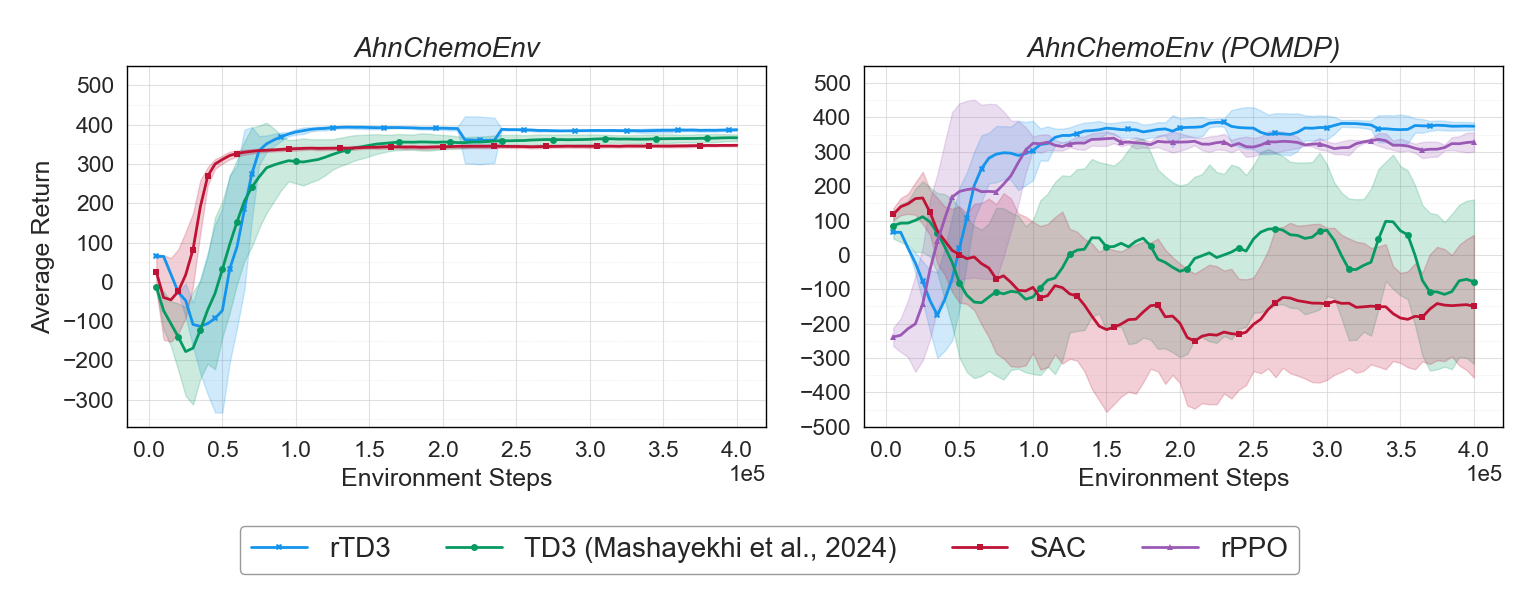}
        \caption{Evaluation performance under full and partial observability. Under full observability, all methods achieve competitive performance, whereas under partial observability, recurrent architectures show stronger and more stable performance.}

    \label{fig:Learning Curve}
\end{figure}
We employ the separate recurrent actor--critic architecture proposed in \cite{ni2022recurrentmodelfreerlstrong}, using independent long short-term memory (LSTM) encoders for the actor and critic. In the actor, the LSTM processes the interaction history and produces a history-dependent latent representation, which is combined with a direct embedding of the current observation to generate the continuous chemotherapy dose. This direct current-observation pathway acts as a shortcut connection; it preserves immediately available information and reduces the burden on the recurrent encoder to memorize the current observation. In our chemotherapy setting, this design helps the agent remain responsive to immediate indicators such as the current tumor burden, while the recurrent history is used primarily to infer the unobserved normal-cell dynamics. Similarly, the critic uses its own LSTM-based history representation together with the current observation--action pair to estimate the twin action-value functions of TD3. This separate design is adopted because it was found to be more stable than a shared recurrent actor--critic encoder\cite{ni2022recurrentmodelfreerlstrong}.
Let \[
\tau_t = (o_0, a_0, r_1, o_1, \dots, a_{t-1}, r_t, o_t)
\]
denote the observable interaction history up to time step \(t\). In the recurrent setting, the policy conditions its action on this history:
\[
a_t = \pi_\theta(\tau_t).
\]
This allows the agent to exploit temporal context when making treatment decisions.
During training, observed trajectories are stored in a replay buffer and sampled as subsequences with a fixed context length of 32 steps. Experience is stored as continuous episodes, from which overlapping subsequences are randomly sampled; hidden states are initialized to zero at the start of each subsequence rather than using an explicit burn-in period, avoiding stale hidden states during off-policy updates and preventing representational inconsistencies arising from the use of stale hidden states sampled from the replay buffer, which could destabilize training. To avoid carrying hidden-state information across independent episodes, sequence batches are masked at episode boundaries. To assess sensitivity to this choice, several context lengths were evaluated; all yielded comparable long-term performance. A 32-step context was therefore retained as a balanced trade-off between temporal coverage and computational efficiency. These sampled sequences are then used to train the recurrent critic through temporal-difference learning and to update the actor policy using the critic's Q-value estimates. A compact TD target can be written as \cite{TD3OriginalPaper,ni2022recurrentmodelfreerlstrong}

\[
y_t = r_t + \gamma (1 - d_t)\min_{i \in \{1,2\}} Q_i'(\tau_{t+1}, a'_{t+1}),
\]
where $Q_i'$ denotes the target critic networks and $a'_{t+1}$ is the action produced by the target actor from the next history. In this way, the recurrent encoder processes past observations together with previous actions and rewards, enabling the policy to capture temporal dependencies and better infer hidden aspects of the underlying system. An overview of the adopted architecture is shown in Figure~\ref{fig:rtd3_clean_ieee}


\begin{figure*}[!t]
    \centering
    \includegraphics[width=\textwidth]{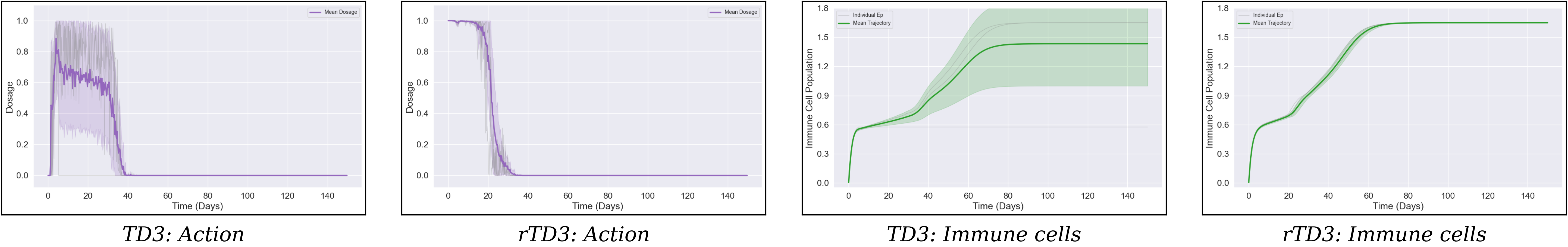}
    \caption{Action and immune-cell trajectories under partial observability over 30 \textit{in silico} episodes. The recurrent policy exhibits more structured dosing behavior and a more stable immune response.}
   \label{fig:traj_action_immune}
\end{figure*}

\section{Results}

We evaluate rTD3, TD3, and SAC on the chemotherapy control task in \textit{AhnChemoEnv} from DTR-Bench \cite{DTRBench} under both fully and partially observable settings. To further validate the benefits of memory under partial observability, we also evaluate rPPO~\cite{schulman2017ppo} exclusively in the POMDP setting, using the implementation from Stable Baselines3~\cite{stable-baselines3}. TD3 serves as the main non-recurrent baseline, as it was previously applied to chemotherapy dose control by Mashayekhi et al. \cite{Mashayekhi2024DRLChemo}. In a partially observable case, the normal-cell population is hidden, observations are noisy, and the underlying state dynamics are subject to stochastic perturbations. Because PK/PD variability introduces additional uncertainty through inter-patient heterogeneity, it is removed here so that the comparison can focus on partial observability and noise while keeping the underlying patient dynamics fixed. All methods are trained and evaluated under the same protocol over 10 shared random seeds. Both the fully observable and partially observable settings are trained for $4 \times 10^5$ environment steps. All experiments were executed on an NVIDIA GeForce RTX 4070 Ti GPU. Evaluation curves report the mean evaluation return across the 10 random shared seeds, computed over 25 evaluation episodes every $5 \times 10^3$ environment steps; shaded regions denote the standard deviation.

\subsection{Learning under Full and Partial Observability}
We first examine how observability affects learning performance across recurrent and non-recurrent methods. Figure~\ref{fig:Learning Curve} shows the evaluation performance of rTD3, TD3, and SAC under full and partial observability, alongside rPPO in the POMDP setting. Under full observability, all methods improve during training and achieve competitive performance, indicating that the task can be handled effectively when the full system state is available. In this setting, the advantage of recurrence is limited, since the policy can rely directly on the current observation. This behavior changes under partial observability, where the agent must act on incomplete and noisy observations. When the normal-cell population is hidden and the observation process is corrupted by noise, the performance gap becomes more pronounced. In the POMDP setting, the recurrent method remains clearly stronger and more stable, whereas TD3 and SAC show degraded performance, large oscillations in evaluation return, substantially higher variance across seeds, and less reliable convergence. In particular, the non-recurrent methods do not improve consistently, and temporary gains are often lost later in training, indicating unstable optimization in the presence of hidden and noisy state information. This suggests that temporal information becomes more valuable when the agent must infer hidden state information from noisy observation histories rather than relying on the current observation alone. Overall, these results indicate that recurrence becomes significantly more beneficial under partial observability, where memory helps compensate for missing and noisy information.

Table~\ref{tab:results} quantifies this trend, showing that the performance difference remains limited under full observability but becomes substantially larger under partial observability, where both recurrent architectures substantially outperform their memoryless counterparts. Among these, rTD3 achieves the strongest and most stable final performance, while rPPO also maintains a positive return with relatively low variance, suggesting that temporal context is the primary driver of robustness in this POMDP task. The extremely large final standard deviations of TD3 and SAC under partial observability therefore reflect a robustness issue rather than a small difference in average performance. Taken together with the oscillatory evaluation curves in Figure~\ref{fig:Learning Curve}, these results indicate strong sensitivity to initialization and training stochasticity in memoryless policies, whereas recurrent architectures remain considerably more stable across seeds. The particularly strong performance of rTD3 highlights the viability of off-policy recurrent methods \cite{Kapturowskietal2019}, which is a critical feature for medical applications where policies must ultimately be learned offline from historical patient records \cite{Komorowskietal2018, Levineetal2020} rather than through active, potentially unsafe exploration \cite{Kumaretal2020}.

\begin{table}[H]
\centering
\footnotesize
\renewcommand{\arraystretch}{1.1}
\caption{ \centering \\ Final performance under full and partial observability \\ (mean $\pm$ std over 10 seeds).}
\label{tab:results}
\begin{tabular}{llcc}
\toprule
\textbf{Architecture} & \textbf{Method} & \textbf{Full Obs} & \textbf{POMDP} \\
\midrule
\multirow{2}{*}{Recurrent}  & rTD3 & \textbf{388.8 $\pm$ 4.6} & \textbf{380.4 $\pm$ 12.2} \\
                            & rPPO & -- & 308.4 $\pm$ 37.0 \\
\midrule
\multirow{2}{*}{Memoryless} & TD3  & 367.7 $\pm$ 7.2 & $-49.8 \pm 252.2$ \\
                            & SAC  & 348.0 $\pm$ 4.4 & $-154.1 \pm 219.7$ \\
\bottomrule
\end{tabular}
\end{table}

\subsection{Trajectory Analysis of Learned Policies}
To gain deeper insight into treatment behavior under partial observability, we examine trajectory-level evaluations of rTD3 and TD3 over 30 \textit{in silico} episodes. These two methods are selected for granular analysis as they represent the primary proposed architecture and the established baseline in the chemotherapy control literature \cite{Mashayekhi2024DRLChemo}, respectively; SAC and rPPO serve primarily to corroborate broader performance trends. In all trajectory plots, solid curves represent the mean response across episodes, shaded regions indicate cross-episode variability, and thin gray trajectories correspond to three representative individual episodes.

Figure~\ref{fig:traj_normal_tumor} highlights the resulting state dynamics in terms of normal-cell and tumor-cell populations. The recurrent policy drives the tumor population toward zero more consistently and with lower dispersion, while also enabling stronger recovery of the normal-cell population. In contrast, the non-recurrent baseline shows greater variability across episodes, and its mean tumor trajectory remains above zero, reflecting delayed suppression and the presence of failed episodes that terminate when the normal-cell population falls below the safety threshold defined in the environment. 

A similar pattern is observed in Figure~\ref{fig:traj_action_immune}, where the recurrent method exhibits more structured dosing behavior together with a stronger and more stable immune-cell response throughout the episodes. This pattern is further supported by Figure~\ref{fig:drug_compare}, which shows that the recurrent policy reaches a higher drug concentration early in treatment, maintains it more consistently across episodes, and then exhibits a more synchronized decline toward zero than TD3. This advantage is also consistent with the structure of the underlying ODE system under noisy partial observability. Although the normal-cell population $N(t)$ is hidden from the agent, it directly affects both the reward and the safety constraint, while also influencing the observed treatment dynamics through its coupling with tumor evolution and drug response. At the same time, the observation process corrupts the measured values of $T(t)$, $I(t)$, and $B(t)$, while state noise perturbs the latent transition dynamics themselves. As a result, the current observation alone is an unreliable basis for treatment decisions; similar observed states may correspond to different levels of hidden normal-cell depletion or recovery depending on recent treatment history. By integrating recent histories of observations, past dosing, and rewards, the recurrent policy can form a better implicit estimate of the latent physiological condition, which helps explain its more reliable tumor suppression together with healthier recovery dynamics under partial observability.


\begin{figure}[t]
    \centering
    \includegraphics[width=\columnwidth]{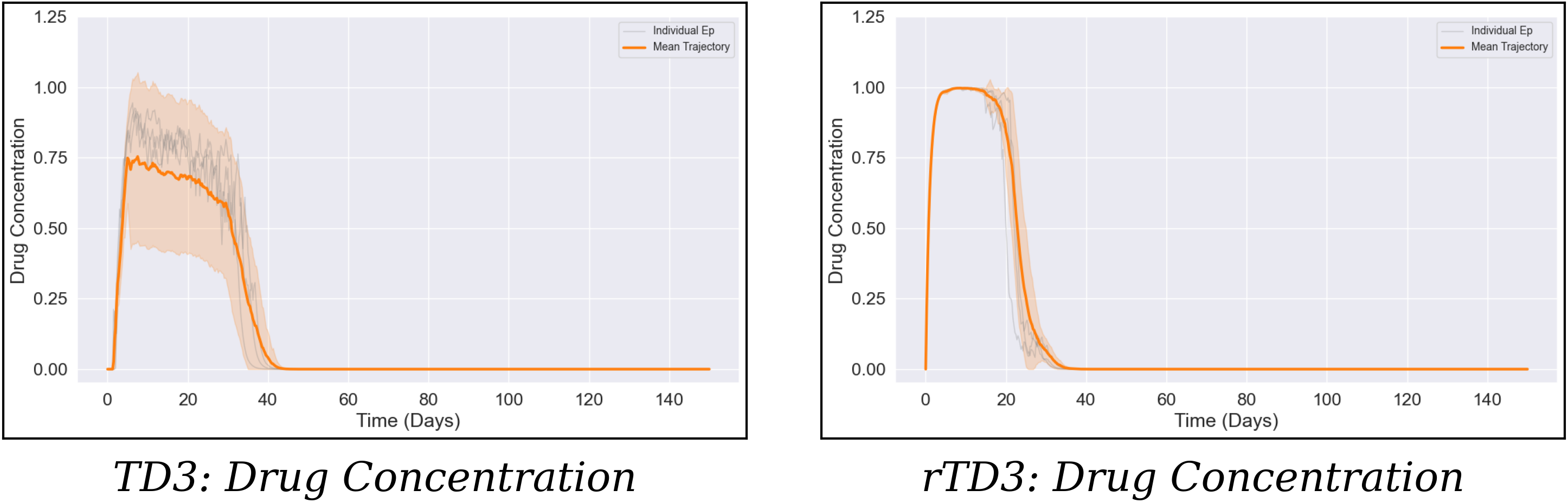}
    \caption{Drug-concentration trajectories under partial observability over 30 \textit{in silico} episodes. The recurrent policy yields higher and more consistent early drug concentration, followed by a more synchronized tapering phase than TD3.}
    \label{fig:drug_compare}
\end{figure}



\section{Conclusion}

This study investigated chemotherapy treatment optimization under partial observability using recurrent actor--critic architectures. While recurrence provides limited benefit in fully observable settings, our results on the \textit{AhnChemoEnv} benchmark demonstrate its necessity under partial observability, where recurrent architectures achieve significantly more stable performance than memoryless baselines, with more consistent tumor suppression and improved normal-cell preservation across seeds. These findings suggest that integrating observation histories allows policies to better infer latent physiological states and make reliable decisions from noisy, incomplete data. Notably, the strong performance of an off-policy recurrent architecture is particularly relevant for clinical applications, where offline learning from historical treatment records is a more realistic deployment scenario than active environment interaction. 

Despite these results, the evaluation remains entirely \textit{in silico}, relying on fixed PK/PD parameters and leaving the approach unvalidated on real patient data. The linear reward scalarization may further inadequately capture the multi-objective trade-offs inherent in oncology, and the reliance on terminal penalties rather than formal safety constraints remains a critical gap for clinical deployment. Future work will therefore incorporate patient-specific PK/PD variability, explicit safety-constrained optimization, and validation on real patient data.

\section*{Acknowledgment}
The authors wish to thank Riccardo De Monte and Marina Ceccon for their insightful discussions and thoughtful feedback, which meaningfully contributed to shaping this work.

\bibliographystyle{IEEEtran}
\bibliography{references}


\end{document}